\documentclass[suppldata]{interact}
\usepackage{placeins}
\usepackage{amsmath,amsfonts}
\usepackage{algorithmic}
\usepackage{algorithm}
\usepackage{array}
\usepackage{booktabs}
\usepackage{xcolor}
\usepackage[caption=false,font=normalsize,labelfont=sf,textfont=sf]{subfig}
\usepackage{textcomp}
\usepackage{stfloats}
\usepackage{booktabs}
\usepackage{url}
\usepackage{hyperref} 
\usepackage{verbatim}
\usepackage{graphicx}
\usepackage{cite}
\usepackage{booktabs}
\usepackage{epstopdf}% To incorporate .eps illustrations using PDFLaTeX, etc.
\usepackage[caption=false]{subfig}% Support for small, `sub' figures and tables
\usepackage[numbers,sort&compress]{natbib}

\usepackage[numbers,sort&compress]{natbib}% Citation support using natbib.sty
\bibpunct[, ]{[}{]}{,}{n}{,}{,}% Citation support using natbib.sty
\theoremstyle{plain}% Theorem-like structures provided by amsthm.sty

\theoremstyle{definition}

\theoremstyle{remark}

\begin{document}

\title{MambaMPD: A Mamba-Driven Segmentation Framework for Marine Pollution Detection from Remote Sensing Imagery}

\author{
\name{Shuaiyu Chen\textsuperscript{a},Wei Han\textsuperscript{b}, Peng Ren\textsuperscript{c}, Chunbo Luo\textsuperscript{a} and Zeyu Fu\textsuperscript{a}\thanks{Corresponding to Zeyu Fu. Email: z.fu@exeter.ac.uk}}
\affil{\textsuperscript{a}Department of Computer Science, University of Exeter, Exeter, United Kingdom;\\ \textsuperscript{b}School of Computer Science, China University of Geosciences, Wuhan, China;\\ \textsuperscript{c}College of Oceanography and Space Informatics, China University of Petroleum (East China), Qingdao, China}
}
\maketitle

\begin{abstract}

Accurate detection of marine pollution is essential for protecting coastal ecosystems and marine biodiversity.
Recently, vision Mamba-based approaches have shown promise in remote sensing semantic segmentation due to their ability to efficiently capture
long-range dependencies and global context. However, their potential remains largely unexplored in the context of Marine Pollution Detection (MPD) with distinct challenges, including low signal-to-noise ratios, spatial fragmentation of pollution patterns, and indistinct boundaries caused by strong visual similarity between pollutants and the surrounding marine environment. To address these challenges, we propose MambaMPD, an enhanced Mamba-based framework tailored for marine pollution detection. MambaMPD incorporates two targeted modules that enhance the Mamba encoder with complementary structural priors: the Frequency-Aware Augmentation (FAA) module and the multi-scale Edge-Guided Attention (EGA) module. The FAA module strengthens the encoding process by integrating Wavelet Transforms, which decompose features into multi-scale frequency subbands. This enables the model to efficiently capture both low-frequency contextual semantics and high-frequency structural details, essential for identifying small, low-contrast, and irregular pollution patterns. Meanwhile, the EGA module adaptively integrates hierarchical Laplacian-derived multi-scale boundary cues into deep semantic representations, guiding the refinement of encoder features before decoding, thereby sharpening boundary delineation and alleviating ambiguity in visually confusing and spatially fragmented marine scenes. Additionally, a U-Net-style decoder equipped with squeeze-and-excitation attention and deep supervision is employed to progressively recover and refine semantic and spatial features across multiple scales. Extensive experiments on two benchmark marine pollution detection datasets demonstrate that the proposed method outperforms the compared
methods in mIoU while maintaining significantly lower
computational cost than foundation-model-based approaches. On MADOS it surpasses OSDMamba by 3.6\% in F1, and on M4D it improves
Oil Spill IoU by 6.82\% over TransOilSeg. The source code of our approach will be available at \url{https://github.com/Multimodal-Intelligence-Lab-MIL/MambaMPD}.

% On MADOS, MambaMPD surpasses OSDMamba by 3.6\% in F1. On M4D, it achieves a particularly notable 6.82\% improvement on the OilSpill category over widely used segmentation architectures, demonstrating strong capability in identifying challenging pollution regions. 

\end{abstract}


\section{Introduction}
\label{intro}
Marine pollution monitoring is a highly demanding yet societally vital task. Pollutants such as oil spills~\cite{dehghani2023oil}, plastic debris~\cite{ma2023global}, and chemical contaminants~\cite{mccarthy2025satellite} pose serious threats to fragile coastal ecosystems and marine biodiversity, while also disrupting navigation and undermining the sustainability of the blue economy~\cite{solberg2007oil,mkrtchyan2018new,varotsos2019new}.
Recent advances in aerospace and sensor technologies have greatly improved access to remote sensing data, including synthetic aperture radar (SAR)\cite{krestenitis2019oil} and multispectral imagery \cite{kikaki2024detecting}. Accurately delineating polluted regions from satellite or aerial imagery is therefore crucial for both rapid emergency response and long-term environmental management~\cite{cernian2025advances,wan2025systematic,chen2025vision}, and contributes directly to the United Nations Sustainable Development Goals, in particular SDG 14, in line with the recognised role of remote sensing in advancing the SDGs~\cite{varotsos2020remote}.
% Accurately delineating polluted regions from satellite or aerial imagery is therefore crucial for both rapid emergency response and long-term environmental management~\cite{cernian2025advances}.

Traditional marine pollution detection methods, such as thresholding~\cite{silliman2016thresholds,hook2020beyond,xu2020oil}, clustering~\cite{yang2015dynamic,capizzi2016clustering}, and Markov Random Field-based segmentation~\cite{moctezuma2014measuring,xu2015oil,lopez2006contextual} primarily rely on hand-crafted features like colour, texture, and spatial consistency. While computationally efficient, these approaches often lack robustness and generalisability in complex marine environments.
Recent progress has seen a shift toward learning-based models, evolving from CNN-based methods~\cite{mahmoud2023oil,li2023ds}, to Transformer-based architectures~\cite{zhong2022nt,liu2023rethinking,zhang2022transformer,chai2025transoilseg}, and more recently, foundation model-based approaches like SAM-OIL~\cite{wu2024compositional}. These developments reflect a transition from task-specific pipelines to prompt-driven, globally-aware segmentation frameworks.
More recently, the Mamba architectures~\cite {gu2023mamba,liu2024vmamba,liu2024swin},  based on state-space models (SSMs)~\cite{gu2023mamba}, have gained attention for their ability to efficiently capture long-range dependencies and model global context. 
While Mamba-based models have shown promise in general remote sensing segmentation~\cite{chen2024rsmamba}, their potential remains largely unexplored in the context of marine pollution detection.

\begin{figure*}[!t]
    \centering
    \includegraphics[width=4in]{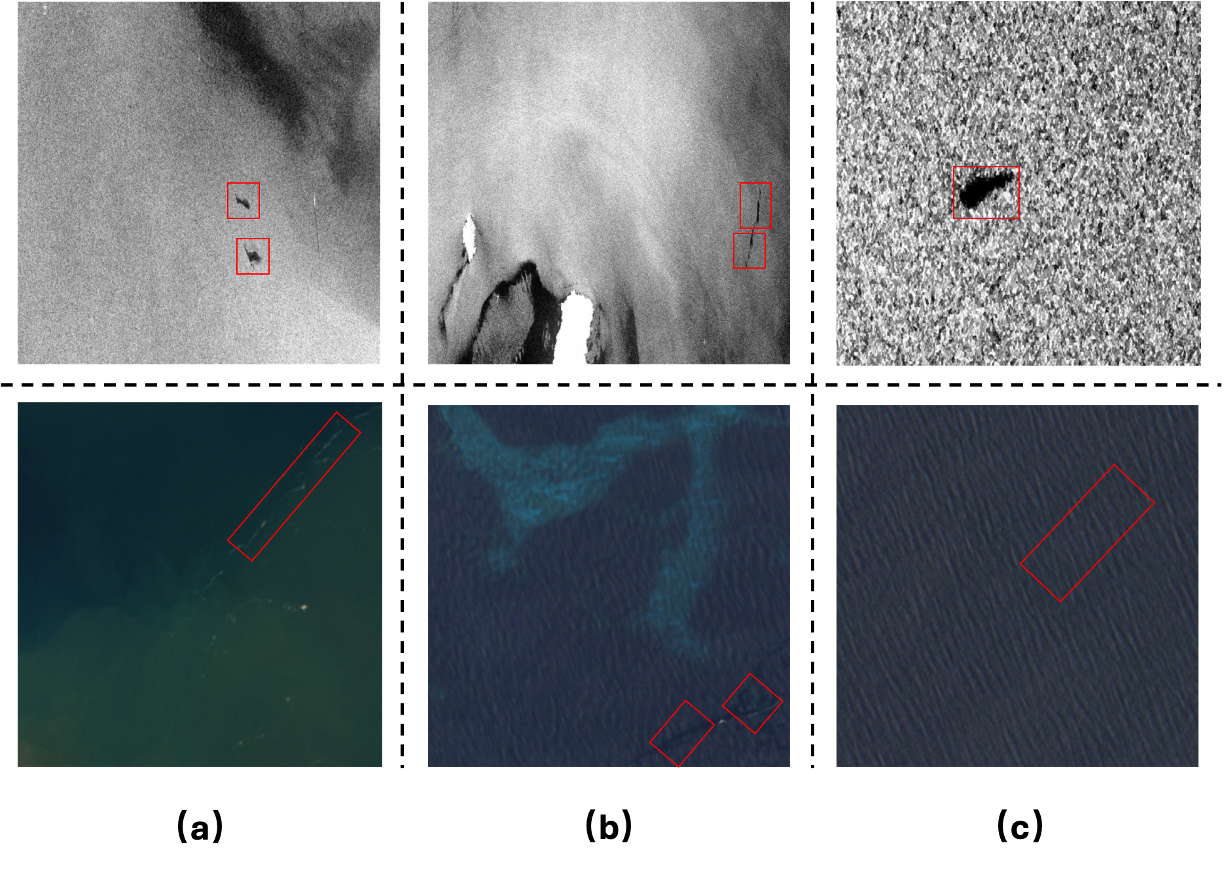}
    \caption{Illustration of key challenges in marine pollution detection across SAR imagery (top row, from M4D dataset) and multispectral imagery (bottom row, from MADOS dataset). (a) Small and dispersed pollution targets (highlighted by red boxes) that are easily obscured by background clutter and noise, making fine-grained detection difficult. (b) Visual and backscatter ambiguity, where oil spills, look-alike dark spots, and natural marine features (e.g., algae, turbid water) exhibit highly similar appearances, leading to frequent misclassification. (c) Indistinct and fragmented pollution boundaries caused by speckle noise in SAR and dynamic sea surface conditions in optical imagery, posing challenges for precise boundary delineation.}
    \label{fig:guang}
\end{figure*}

Despite significant progress in learning-based methods for marine pollution detection, several fundamental challenges persist across both multispectral and SAR imagery, as illustrated in Fig.\ref{fig:guang}. These challenges are not specific to any particular architecture but remain largely unresolved across existing CNN-based, Transformer-based, and state-space model-based approaches:

1)Pollution targets such as oil spills and plastic patches typically manifest as small, dispersed regions, often captured by medium-resolution sensors like Sentinel-2 and SAR platforms \cite{kikaki2024detecting}. These modalities lack the spatial or textural granularity required to capture fine structures, making models highly sensitive to noise~\cite{al2020sensors}.

2)Pollutants (such as emulsified oils, rainbow sheens, and plastics) often share similar visual or backscatter patterns with non-pollutant features such as seawater, ship wakes, plankton, and algae~\cite{sannigrahi2022development,gomez2022learning}, leading to substantial semantic ambiguity and misclassification.~\cite{heiselberg2020ship}.

3)Dynamic sea surface conditions, driven by waves, currents, and biological activity, further obscure the boundaries between pollution and background regions~\cite{liu2015survey}.

\begin{figure*}[!t]
    \centering
    \includegraphics[width=5in]{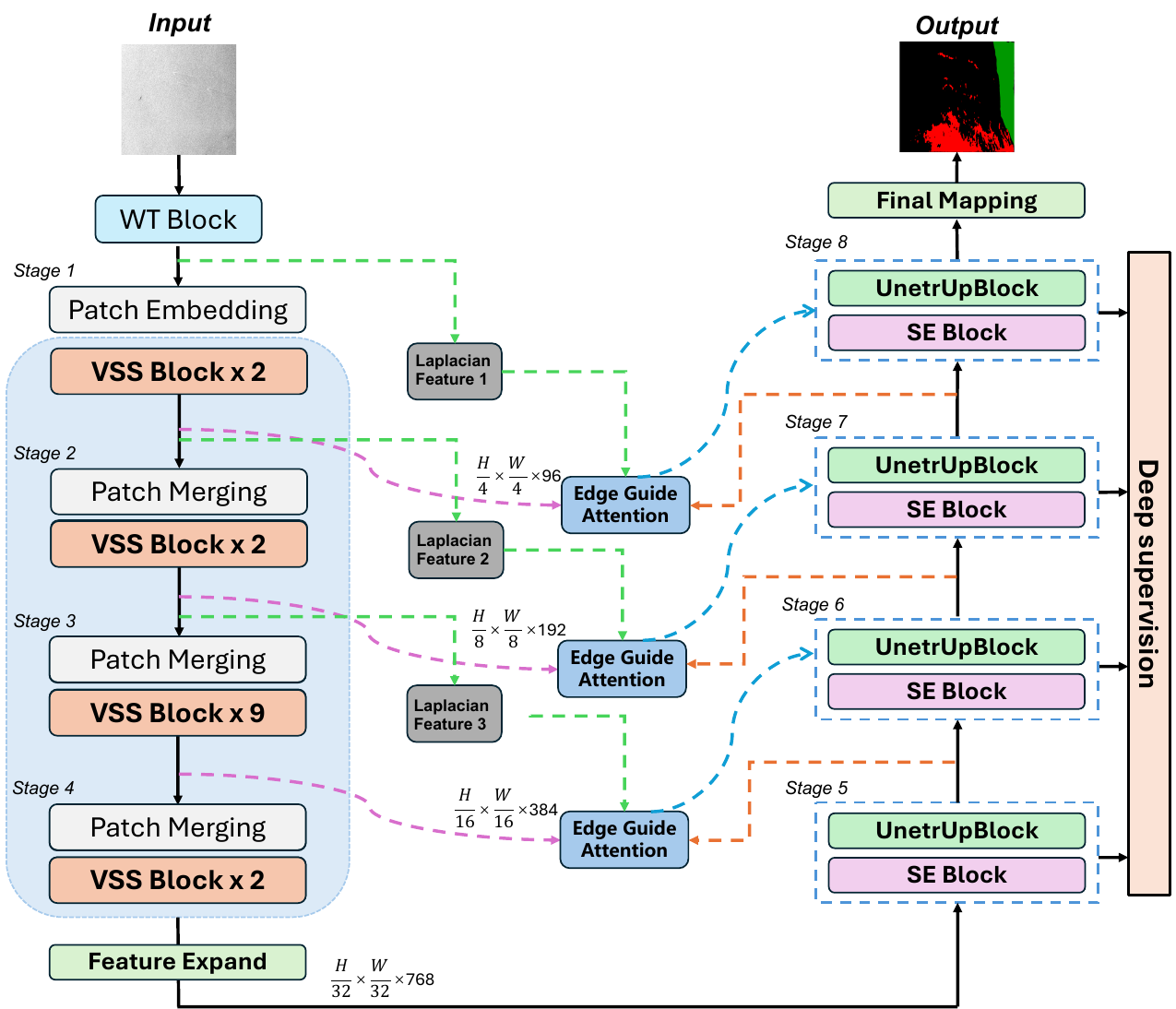}
    \caption{Overall architecture of the proposed framework. The encoder extracts multi-scale features from the input image through a wavelet-enhanced hierarchical backbone consisting of a WT block, patch embedding, patch merging, and stacked VSS blocks. To strengthen boundary modeling, Laplacian features from different stages are incorporated into edge-guided attention modules, which refine encoder representations before they are passed to the decoder. The decoder progressively restores spatial resolution via stacked UnetrUpBlocks and SE blocks, where skip connections explicitly transfer encoder features to the corresponding decoder stages for structural detail recovery. Deep supervision is further applied to intermediate decoding layers, and the final prediction is generated through the output mapping layer.}
    \label{fig:stru}
\end{figure*}

While Mamba-based architectures have demonstrated strong capability in capturing long-range dependencies with linear complexity, they still lack dedicated mechanisms to address the above challenges in the context of marine pollution detection. A recent work, OSDMamba~\cite{OSDMamba}, adapts Mamba to SAR oil-spill
segmentation but retains a standard VSS encoder, and therefore offers no
mechanism for the high-frequency cues of small, low-contrast targets or for boundary preservation under speckle and dynamic
sea states; it is moreover confined to binary,
single-modality detection. To this end, we propose MambaMPD, a Mamba-driven segmentation framework tailored for marine pollution detection, as shown in Fig. \ref{fig:stru} (a). Built upon the expressive capacity of VMamba~\cite{liu2024vmamba}, MambaMPD incorporates several targeted adaptations to address the unique challenges posed by marine pollutants across both SAR and multispectral imagery.
A key component of MambaMPD is the Frequency-Aware Augmentation (FAA) Module, which mainly integrates Wavelet Transforms into the Mamba encoder to decompose features into multi-scale frequency subbands. This allows the model to efficiently capture both contextual semantics and fine structural details, enhancing its ability to detect small, low-contrast, and irregular marine pollution targets.

Another key component of MambaMPD is the multi-scale Edge-Guided Attention (EGA) module, which adaptively integrates Laplacian-based edge cues with semantic features extracted at multiple encoder depths. EGA is specifically designed to enhance boundary preservation under conditions of spatial sparsity. Additionally, a U-Net-style decoder equipped with Squeeze-and-Excitation (SE) attention and deep supervision is employed to progressively recover and refine semantic and spatial features across multiple scales.
To validate the effectiveness of the proposed MambaMPD framework, we conduct comprehensive experiments on two challenging marine pollution detection datasets, M4D~\cite{krestenitis2019oil} and MADOS~\cite{kikaki2024detecting}, featuring realistic sea conditions, ambiguous boundaries, and pollutant-background similarities. Following Kikaki et al.~\cite{kikaki2024detecting}, our experiments focus on oil spills and
marine debris, the two major marine pollutants observable in satellite
imagery; extension to other pollution types is discussed in Section~6.

In summary, our contributions are shown as follows:

1) We propose MambaMPD tailored for marine pollution detection, addressing domain-specific challenges in SAR and multispectral imagery. 

2) We introduce an FAA module, which augments the Mamba encoder by incorporating WT to capture both low-frequency contextual semantics and high-frequency structural cues, crucial for detecting small, low-contrast, and irregular pollution targets, without introducing noticeable parameter overhead.

3) We introduce a multi-scale EGA module to effectively fuse multi-scale edge features with deep semantic representations. This module improves boundary localisation and reduces confusion in visually similar marine pollution classes.

4) Experimental results show that the proposed MambaMPD achieves the best mIoU among compared methods, outperforming TransOilSeg by 6.82\% in Oil Spill
IoU on M4D and MariNeXt by 4.2\% in F1-score on MADOS.

\section{Related Work}

\subsection{Marine Pollution Detection from Remote Sensing}
Early approaches to marine pollution detection in remote sensing primarily relied on expert-defined heuristics and rule-based strategies. These traditional methods~\cite{heiselberg2020ship,sannigrahi2022development,gomez2022learning,varotsos2018pollution,varotsos2020novel}, such as spectral thresholding, morphological filtering, and geometric descriptors, were typically tailored to detect a single pollutant type. However, they struggle in multi-class scenarios due to limited adaptability to competing marine elements.

For instance, Kikaki et al.~\cite{kikaki2024detecting} observed that rule-based systems often confuse plastic debris with natural water patterns and misclassify clean water regions when appearance overlap is high.
Similar issues have been reported in optical imagery, where marine mucilage is mistaken for floating waste~\cite{hu2022spectral}, and in SAR data where fine debris is difficult to distinguish from ships or sea clutter due to similar radar backscatter~\cite{qi2022capacity}. While fast and interpretable, these methods’ reliance on low-level cues and rigid rules limits their scalability and generalisation in real-world applications.

To address these shortcomings, conventional machine learning (ML) techniques have been explored, where handcrafted features, such as texture, shape, or contextual descriptors, are paired with classifiers like support vector machines or random forests~\cite{kikaki2022marida,mifdal2021towards,mikeli2022sentinel}. These approaches introduce greater adaptability and can incorporate spatial patterns to detect complex marine targets. In particular, ML models have shown promise in recognising small-scale or texture-rich pollutants under varying conditions~\cite{hu2022spectral}. Nevertheless, their success is often constrained by the quality of manually extracted features and the inherent imbalance in marine datasets. Moreover, specific pollutant types, such as large plastic patches or dispersed oil films, present sporadic appearances and irregular shapes~\cite{dierssen2019hyperspectral,garaba2018airborne}, further challenging the generalisation capabilities of these models.

With the rise of deep learning, semantic segmentation techniques have been increasingly adopted in remote sensing applications. Classic encoder–decoder architectures such as U-Net~\cite{ronneberger2015u} and DeepLab~\cite{deeplabv3} extract hierarchical features and leverage skip connections and atrous convolutions to maintain both global semantic context and fine-grained spatial details~\cite{li2024transformer,kotaridis2021remote,dey2010review}. These approaches have been progressively extended to the task of marine pollution detection.
A series of early works focused on oil spill detection using SAR imagery as a binary segmentation task. For example, Mahmoud et al.~\cite{mahmoud2023oil} employed a standard U-Net to separate oil spills from background water, achieving promising results under simple scenarios. However, this setup lacks scalability for more complex or diverse marine scenes. To address blurry boundaries and noisy textures in SAR imagery, CBD-Net~\cite{9568691} introduced contextual and boundary supervision, enhancing small target segmentation and achieving state-of-the-art performance on the SOS dataset.
Further improving SAR-based segmentation, DGNet~\cite{10029889} incorporated the intrinsic distribution of SAR backscatter values into a latent variable inference generation loop, demonstrating data-efficient training and superior generalisation with limited annotations. Similarly, SRCNet~\cite{10683756} designed a competing dual-network structure leveraging a SAR-specific image representation, which enhanced learning from small-scale datasets and achieved accurate segmentation under constrained supervision.

Beyond binary segmentation, recent works began addressing the need to distinguish between multiple types of marine targets in SAR images, such as ship wakes, look-alike dark spots, and other oil-like artefacts. The M4D dataset~\cite{krestenitis2019oil} represents an effort in this direction, offering a richer annotation schema for evaluating fine-grained pollutant detection. 
Building on this, Wu et al.~\cite{wu2024compositional} proposed SAM-OIL, leveraging vision foundation models for few-shot and interactive segmentation. However, its performance is constrained by prompt quality and the absence of explicit structural cues, limiting precision in noisy, cluttered marine scenes.
Chai et al.~\cite{chai2025transoilseg} proposed TransOilSeg, which performs well on large-scale imagery with clear oil slicks but may struggle in scenes with small, fragmented, or low-contrast targets. 

To extend beyond SAR, Duarte et al.~\cite{duarte2023automatic} developed a deep learning pipeline for detecting marine debris in multispectral imagery. Kikaki et al.~\cite{kikaki2024detecting} introduced the MADOS dataset and proposed MariNeXt, a CNN-based model that fuses spectral and spatial cues. While effective across pollutant types and sensors, MariNeXt lacks mechanisms to capture structural or edge information crucial for complex scenes. 

In summary, existing models still struggle with key challenges such as small object size, low contrast and boundary confusion~\cite{garcia2017review,hao2020brief,minaee2021image}. Distinct from prior works, our proposed MambaMPD explicitly addresses these limitations through the introduction of the Frequency-Aware Augmentation (FAA) module and Edge-Guided Attention (EGA) module, which enhance structural sensitivity, preserve contextual semantics, and improve the delineation of fine-grained object boundaries.

\subsection{Vision Mamba in Remote Sensing Segmentation}
A growing body of research has explored the adaptation of the Mamba architecture~\cite{gu2023mamba} to remote sensing image segmentation, showcasing its efficiency in modeling long-range dependencies for high-resolution and large-scale spatial tasks. For instance, RS-Mamba~\cite{zhao2024rsmamba} introduces an omnidirectional selective scan to process gigapixel-scale remote sensing images with linear computational cost, achieving competitive results in land cover segmentation and change detection. RS3Mamba~\cite{ma2024rs3mamba} complements a convolutional backbone with a state-space auxiliary branch, enabling effective multi-level feature fusion for urban semantic segmentation. In the bi-temporal domain, ChangeMamba~\cite{chen2024changemamba} employs a Vision-Mamba encoder and a set of specialized decoders to capture temporal correlations, surpassing traditional CNN and Transformer baselines on several change detection benchmarks. Meanwhile, RSMamba~\cite{chen2024rsmamba} further demonstrates the potential of Mamba in spatial classification by introducing dynamic multi-path activation to remove causal constraints and better adapt to 2D spatial tasks. Some preliminary segmentation works, such as in~\cite{zhao2024rsmamba, chen2024mambau}, have verified the feasibility of Mamba in land cover mapping and building footprint extraction.

Despite their promising capabilities, these efforts concentrate on
land-oriented tasks such as land cover mapping, urban segmentation, and
change detection; marine pollution detection, which demands sharper
boundary delineation and sensitivity to high-frequency structural
details for small, sparse, and visually ambiguous pollutants, has
received little attention. Among Mamba-based approaches, OSDMamba~\cite{OSDMamba}
is the closest to our work, applying a standard VSS encoder to binary
oil-spill segmentation in SAR imagery. MambaMPD differs in three
respects: it augments the encoder with frequency-aware decomposition
(FAA) and injects multi-scale edge priors into decoding (EGA); it
extends the task from binary oil-spill segmentation to multi-class
marine pollution detection; and it is validated on both SAR and
multispectral data. A quantitative comparison controlling for model
capacity is given in Section~4.4.2.
\begin{figure*}
    \centering
    \includegraphics[width=5.5in]{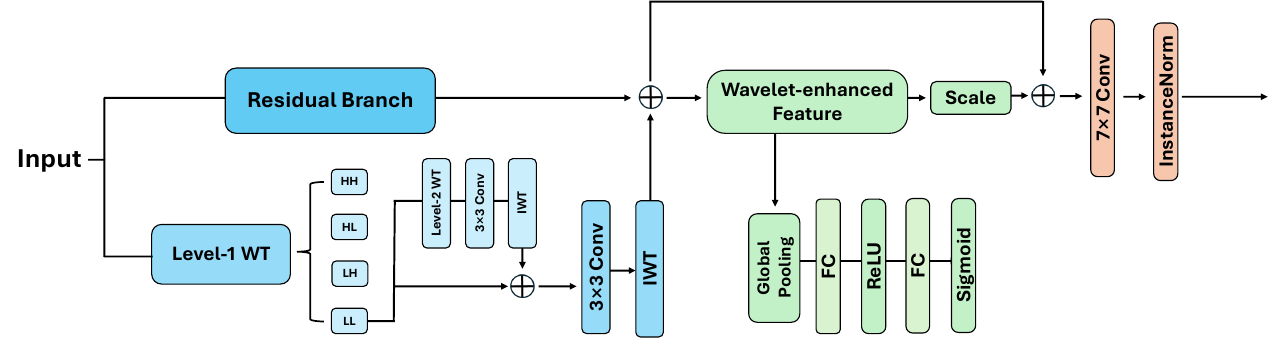}
    \caption{Illustration of the proposed Frequency-Aware Augmentation Module (FAA). Given an input feature, FAA employs a dual-branch design consisting of a residual branch and a wavelet transform branch. The wavelet branch performs level-1 wavelet decomposition to separate low- and high-frequency components, while the high-frequency information is further refined through an additional wavelet-based convolution operation. The processed wavelet features are fused with the residual features and subsequently recalibrated by a channel-attention mechanism to emphasize informative frequency responses. The final enhanced representation is produced through residual aggregation, followed by a $7 \times 7$ convolution and instance normalization.}
    \label{fig:stru2}
\end{figure*}

\section{METHOD}
 
Marine pollution detection from remote sensing imagery can be formulated as a pixel-wise multiclass semantic segmentation task. Given an input image $X \in \mathbb{R}^{H \times W \times C_{\text{in}}}$, the goal is to learn a mapping $Y = \mathcal{F}_\theta(X) \in \mathbb{R}^{C \times H \times W}$ that assigns each pixel a probability distribution over $C$ semantic categories. In our formulation, $X$ can represent either multispectral optical images or single-/dual-polarization SAR data, enabling the framework to operate in a modality-agnostic manner. The following subsections present the proposed MambaMPD framework in detail.
 
\subsection{Mamba-based Encoder}
 
MambaMPD adopts the Visual State Space (VSS) block~\cite{liu2024vmamba} as its fundamental building unit. The VSS block replaces traditional attention mechanisms with selective state-space dynamics, efficiently modeling long-range dependencies with linear complexity. Given a sequence of visual tokens $x \in \mathbb{R}^{H \times W \times C}$, the core state-space recurrence is defined as:
\begin{equation}
s_t = A \cdot s_{t-1} + B \cdot \text{Conv}_{\text{proj}}(x_t), \qquad y_t = \sigma(G_t) \odot (C \cdot s_t),
\end{equation}
where $A$ and $B$ are learnable transition matrices governing the system dynamics, $G_t$ is a learned spatial gate, and $C$ is a readout matrix. The output is restored to the original spatial dimension via a convolutional projection $\hat{x}_t = \text{Conv}_{\text{out}}(y_t)$. To capture spatial context from multiple orientations, the VSS block further employs a two-dimensional selective scanning (SS2D) mechanism~\cite{liu2024swin} that unfolds the feature map along four directional paths and merges the resulting representations.
 
\begin{figure}
    \centering
    \includegraphics[width=3in]{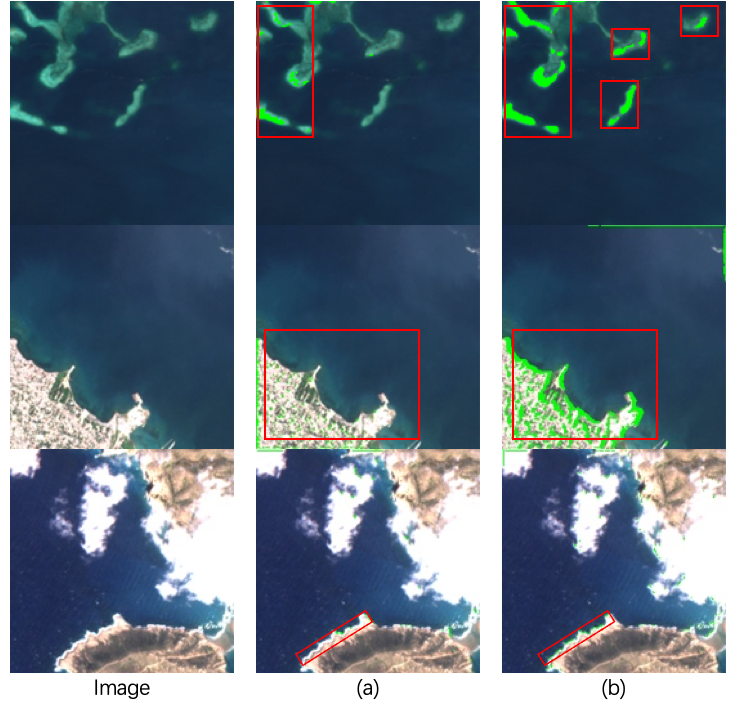}
    \caption{(a) shows the feature response using the baseline Mamba encoder, while (b) demonstrates the enhanced performance achieved by integrating FAA into the baseline Mamba encoder. 
    Red boxes highlight regions where the FAA enhances feature responses in small targets and boundary regions, which are critical for detecting sparse marine pollutants in complex backgrounds.}
    \label{fig:WTConv}
\end{figure}

Here, $s_t \in \mathbb{R}^{N}$ denotes the hidden state
of dimension $N$, $x_t \in \mathbb{R}^{C}$ is the feature vector of
a single visual token, and $A \in \mathbb{R}^{N \times N}$,
$B \in \mathbb{R}^{N \times C}$, $C \in \mathbb{R}^{C \times N}$ are
the state-space matrices. To apply this 1-D recurrence to 2-D
feature maps, SS2D~\cite{liu2024vmamba} unfolds each
$H \times W$ feature map along four directional paths
(left-to-right, right-to-left, top-to-bottom, bottom-to-top)
to produce four sequences of length $H \times W$. Each sequence
is processed independently; the four outputs are then summed
element-wise and reshaped to $H \times W \times C$. The overall encoder follows the hierarchical design of VMamba-Tiny~\cite{liu2024vmamba} and Swin-UMamba~\cite{liu2024swin}, performing $2\times$ spatial downsampling at each stage. The four encoder stages are composed of \{2, 2, 9, 2\} VSS blocks, respectively, providing a balanced trade-off between depth and efficiency. However, as shown in Fig.~\ref{fig:WTConv}(a), our empirical observations reveal that the standard Mamba-based encoder suffers from limited frequency sensitivity and poor edge preservation, particularly for small, sparsely distributed marine pollutants under complex backgrounds.
 
\subsection{Frequency-Aware Augmentation Module}

Frequency decomposition is motivated by the physics
of marine pollution imagery. In SAR, oil spills dampen capillary
and short gravity waves, producing smooth dark patches; the
large-scale backscatter contrast is carried by low-frequency
components, while the sharp gradient at slick boundaries falls
in high-frequency components~\cite{alpers2017oil}. In multispectral data, pollutants
differ from clean water subtly at a global scale but abruptly
at their edges. Wavelet decomposition separates these two cues:
low-frequency sub-bands encode region-level semantics and
high-frequency sub-bands preserve boundary detail~\cite{liu2018multi}. As shown in
Fig.~\ref{fig:stru}, the FAA module sits before the
patch embedding layer and operates on the raw input
$X \in \mathbb{R}^{H \times W \times C_{\mathrm{in}}}$.
Placing it this early injects frequency cues before spatial
downsampling, so high-frequency detail is not lost to
successive patch merging. To address the limited frequency sensitivity of the standard encoder, we propose a Frequency-Aware Augmentation (FAA) module, as illustrated in Fig.~\ref{fig:stru2}. FAA introduces hierarchical wavelet decomposition together with lightweight convolutional refinement, enabling the encoder to simultaneously capture global contextual structures and fine-grained boundary details that are critical for marine pollution segmentation.
 
Given an input feature tensor $X \in \mathbb{R}^{C \times H \times W}$, we first apply a separable 2D Haar wavelet transform to decompose $X$ into four frequency sub-bands:
\begin{equation}
[X_{LL}, X_{LH}, X_{HL}, X_{HH}] = \mathrm{WT}(X),
\end{equation}
where $X_{LL}$ denotes the low-frequency approximation component, and $X_{LH}$, $X_{HL}$, $X_{HH}$ capture high-frequency details along the horizontal, vertical, and diagonal directions, respectively. Each sub-band is refined by a shared $3 \times 3$ convolutional operator $\phi(\cdot)$ and reconstructed via the inverse wavelet transform (IWT).
 
To further enlarge the effective receptive field and strengthen multi-scale frequency perception, we adopt a hierarchical decomposition strategy where only the low-frequency approximation branch from the previous level is further decomposed:
\begin{equation}
[X_{LL}^{(i)}, X_{LH}^{(i)}, X_{HL}^{(i)}, X_{HH}^{(i)}] = \mathrm{WT}(X_{LL}^{(i-1)}).
\end{equation}
This enables the model to progressively capture increasingly global structures while preserving high-frequency details at each level. The features are then reconstructed in a coarse-to-fine manner:
\begin{equation}
Z^{(i)} = \mathrm{IWT}\big(y_{LL}^{(i)} + \uparrow(Z^{(i+1)}),\; y_{LH}^{(i)},\; y_{HL}^{(i)},\; y_{HH}^{(i)}\big),
\end{equation}
where $y_{LL}^{(i)}$, $y_{LH}^{(i)}$, $y_{HL}^{(i)}$, and $y_{HH}^{(i)}$ are the refined sub-band features at level $i$, $\uparrow(\cdot)$ denotes upsampling to match the spatial resolution of the current level, and $Z^{(i+1)}$ is the reconstructed output from the next coarser level. Through this top-down reconstruction scheme, FAA progressively integrates coarse contextual information with fine structural details across decomposition levels.
 
The DWT and IWT use fixed Haar wavelet
filters implemented as stride-2 convolutions and transposed
convolutions, respectively; their weights are frozen and
receive no gradient updates. The only trainable components
in FAA are the $3 \times 3$ refinement convolutions
$\phi(\cdot)$ applied to each sub-band and the subsequent
SE channel attention block. Gradients pass through the fixed
wavelet layers via the standard chain rule, as through any
non-learnable linear operation.
Table~\ref{tab:complexity} confirms this: FAA adds few
trainable parameters. The reconstructed feature is then fused with a residual branch and passed through a squeeze-and-excitation (SE) block to adaptively recalibrate channel responses. Finally, a $7 \times 7$ depthwise convolution followed by instance normalization is applied to enhance local-global interaction and generate the final augmented feature. 
As illustrated in Fig.~4(b), incorporating FAA into the encoder
leads to more discriminative responses in small-target and
boundary regions. The ablation study
(Table~\ref{tab:ablation}) confirms this quantitatively: FAA
alone improves Oil Spill IoU by \textbf{+5.29\%} over the
baseline, which are particularly important for detecting sparse marine pollutants under complex backgrounds. Notably, as reported in Table~\ref{tab:complexity}, this additional cost is negligible in practice. As reported in Table~\ref{tab:complexity}, FAA adds only 0.30\,M
trainable parameters while providing the
largest single-module gain in Oil Spill IoU.
 
\begin{figure}
\centering
\includegraphics[width=3in]{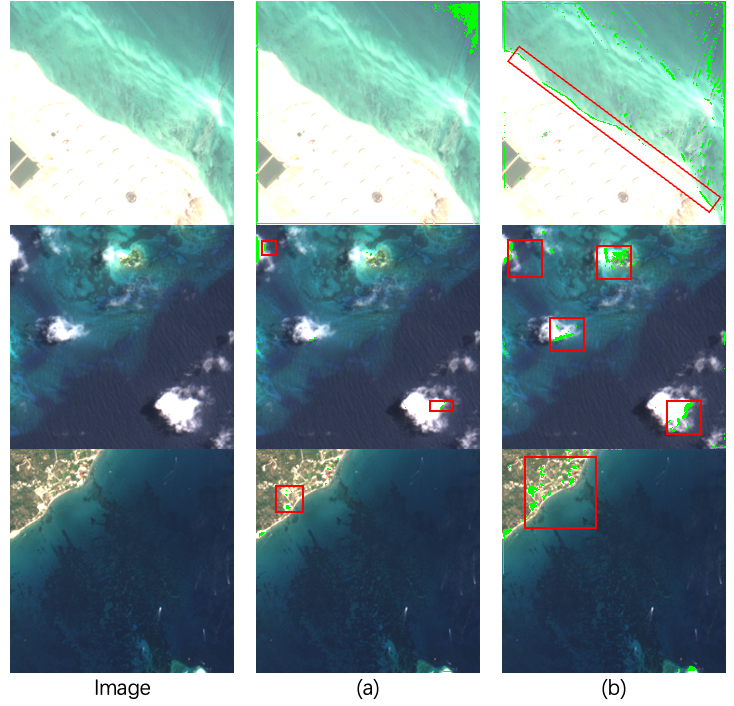}
\caption{(a) presents the baseline feature response, and (b) shows the enhanced feature response with the SE-ResDecoder. Red boxes show that our approach produces more accurate boundaries and consistent segmentation masks, demonstrating improved robustness in small-scale and ambiguous marine pollutant regions.}
\label{fig:SEDec}
\end{figure}

\subsection{Decoder}
 
To improve feature recalibration and semantic refinement during decoding, we adopt a Squeeze-and-Excitation Residual Decoder (SE-ResDecoder), as shown in Fig.~\ref{fig:stru}. Each decoder stage consists of a UnetrUpBlock followed by an SE block. The UnetrUpBlock first upsamples the coarser decoder feature via transpose convolution and concatenates it with the corresponding encoder skip feature. The fused representation is then refined through two residual convolutional blocks composed of convolution, normalization, and LeakyReLU activation. This design enables the decoder to progressively align semantic information from coarse decoder features with structural details from shallow encoder features. Subsequently, a squeeze-and-excitation block performs channel-wise recalibration to emphasize the most informative feature channels.
 
In addition, deep supervision is applied at each decoder stage: a $1 \times 1$ convolution generates an auxiliary segmentation prediction that is upsampled to the original resolution and supervised by the ground-truth mask, facilitating stable gradient propagation and encouraging consistent semantic recovery across decoder stages. As illustrated in Fig.~\ref{fig:SEDec}(a) and Fig.~\ref{fig:SEDec}(b), the proposed SE-ResDecoder enables the network to recover more semantically coherent and spatially precise predictions, particularly in challenging regions with ambiguous boundaries, fragmented structures, and small targets.
 
\subsection{Edge-Guided Attention (EGA) Module}
 
\begin{figure}
    \centering
    \includegraphics[width=3in]{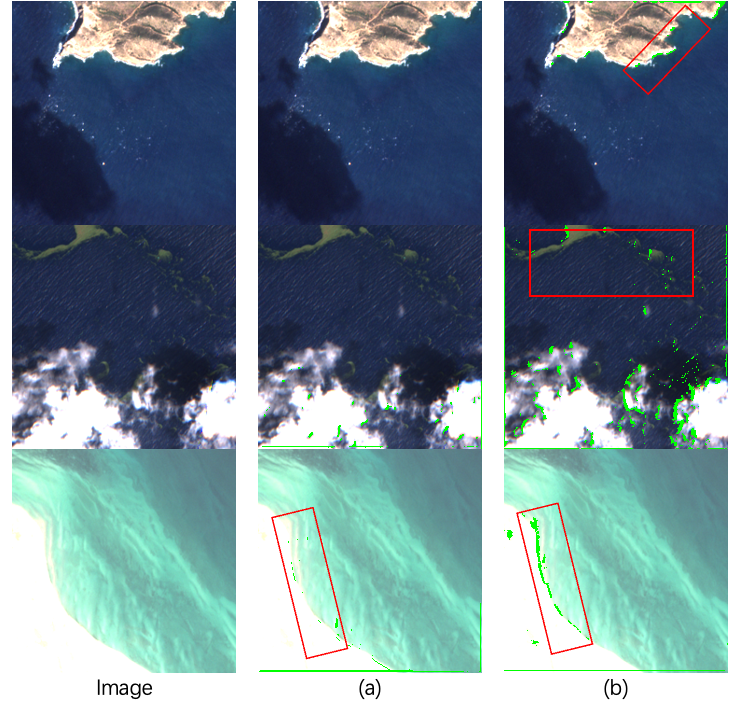}
    \caption{(a) represents the baseline model without edge guidance and (b) represents the baseline model with the proposed EGA module. Red boxes highlight that the baseline with the EGA module produces sharper boundaries and more coherent feature responses.}
    \label{fig:EGAab}
\end{figure}

\begin{figure*}[t]
    \centering
    \includegraphics[width=5.5in]{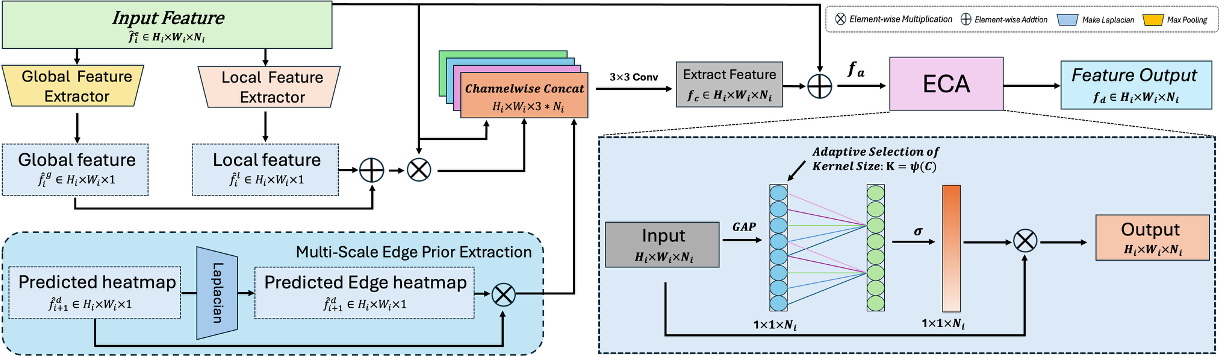}
    \caption{Conceptual overview of the proposed Edge-Guided Attention (EGA) module. The encoder feature is processed by parallel global and local feature extractors to capture complementary contextual and detailed spatial cues. In parallel, the decoder prediction is transformed into an edge-aware guidance map, which is used to modulate the feature representation. The modulated features are then concatenated and refined, followed by channel-wise recalibration using the Efficient Channel Attention (ECA) module, to produce the final edge-enhanced output. This figure is intended as a schematic illustration of the overall interaction mechanism, while the detailed implementation and mathematical formulation are provided in Eqs.~(5)-(7).}
    \label{fig:EA}
\end{figure*}

Although the Mamba-based encoder--decoder architecture exhibits strong global contextual modeling capabilities, Fig.~\ref{fig:EGAab}(a) shows that purely semantic decoding often struggles to recover fine-scale pollution structures, particularly in low-contrast or visually ambiguous ocean regions. These failure cases suggest that decoder features alone are insufficient to provide the geometric cues required for accurate boundary reconstruction. To address this limitation, we introduce an Edge-Guided Attention (EGA) module, as schematically illustrated in Fig.~\ref{fig:EA}. The purpose of EGA is to inject structural guidance into the decoding process by jointly exploiting global context, local structural cues, and edge-aware guidance derived from the decoder prediction.
 
Let $\hat{f}^{e}_{i} \in \mathbb{R}^{H_i \times W_i \times C_i}$ denote the encoder feature at stage $i$, and let $\hat{f}^{d}_{i+1} \in \mathbb{R}^{H_i \times W_i \times 1}$ denote the decoder-side prediction aligned to the same spatial resolution. $\hat{f}^d_{i+1}$ is an intermediate auxiliary prediction
from the deep supervision branch at decoder stage $i{+}1$.
As described in Section~3.3, each decoder stage produces a
single-channel prediction via a $1 \times 1$ convolution and
sigmoid activation. This prediction is bilinearly
interpolated to the resolution of the encoder feature
$\hat{f}^e_i$ before entering EGA. Using these stage-wise
predictions as edge guidance lets EGA access semantic cues
that sharpen as decoding progresses, rather than depending
on a single fixed output. Following the upper-left branch of Fig.~\ref{fig:EA}, we first derive complementary global and local spatial response maps $f_i^{g}$ and $f_i^{loc}$ from the encoder feature via a Global Feature Extractor and a Local Feature Extractor, respectively, where $f_i^{g}$ captures broad contextual saliency and $f_i^{loc}$ preserves local structural details. To explicitly inject boundary cues, the decoder prediction is transformed into an edge-aware map through a Laplacian operator:
\begin{equation}
% f_i^{edge} = \left| \mathcal{L} * \hat{f}^{d}_{i+1} \right| , 
% \mathbf{L} = \begin{bmatrix} 0 & -1 & 0 \\ -1 & 4 & -1 \\ 0 & -1 & 0 \end{bmatrix}
f_i^{\mathrm{edge}}=\mathrm{Norm}\!\left(\frac{1}{3}\!\sum_{k\in\{1,2,4\}}\!\uparrow\!\Big(\big|\,\mathbf{L}*\mathrm{MaxPool}(\hat f_{i+1}^{d};k)\big|\Big)\right),\quad
\mathbf{L}=\begin{bmatrix}0&-1&0\\-1&4&-1\\0&-1&0\end{bmatrix}
\end{equation}

The global and local responses are combined and modulated by the edge prior to form a structurally guided attention map:
\begin{equation}
f_i^{m} = (f_i^{g} + f_i^{loc}) \odot f_i^{edge},
\end{equation}
where $\odot$ denotes element-wise multiplication. The resulting map encodes structurally guided spatial attention informed by both context and predicted boundaries.

This modulated map, together with the global and local response maps, is concatenated with the encoder feature, refined by a $3\times3$ convolution, and combined with the original encoder feature through a residual connection to preserve the original representation while injecting guided spatial emphasis:
\begin{equation}
% f_i^{a} = \operatorname{Conv}_{3\times3}\!\big(\operatorname{Concat}[\hat{f}^{e}_{i},\, f_i^{m}]\big) + \hat{f}^{e}_{i}.
f_i^{c} = \mathrm{Conv}_{3\times3}\!\left( \mathrm{Concat}[\hat{f}_i^{e},\, f_i^{g},\, f_i^{\mathrm{loc}},\, f_i^{m}] \right) + \hat{f}_i^{e}.
\end{equation}
 
Finally, an Efficient Channel Attention (ECA) block~\cite{wang2020eca} is applied to adaptively emphasize informative channels. The ECA block computes a channel descriptor via global average pooling and obtains attention weights through a lightweight 1D convolution, producing the edge-enhanced output $f_i^{a}$ for subsequent decoding. A detailed algorithmic summary of the full EGA pipeline, including multi-scale edge prior construction and cross-scale feature fusion, is provided in Table~\ref{tab:wei}. As shown in Fig.~\ref{fig:EGAab}(b), the proposed EGA module substantially improves boundary fidelity, reduces false positives in ambiguous regions, and facilitates coherent segmentation of thin, irregular, and highly fragmented pollution structures.

\begin{table}[!t]
\centering
\caption{Pseudo-code of the proposed Edge-Guided Attention (EGA) at decoder stage $i$.}
\label{tab:wei}
\footnotesize
\setlength{\tabcolsep}{3pt}
\renewcommand{\arraystretch}{1.05}
\begin{tabular}{p{0.95\linewidth}}
\toprule
\textbf{Input:} Encoder feature $f_i^e \in \mathbb{R}^{C_i \times H_i \times W_i}$, decoder prediction $f_{i+1}^d \in \mathbb{R}^{1 \times H_i \times W_i}$ \\
\textbf{Output:} Edge-enhanced encoder feature $f_i^a \in \mathbb{R}^{C_i \times H_i \times W_i}$ \\
\midrule
\textbf{Phase I: Multi-scale edge prior construction} \\
1. Fine-scale Laplacian response: $E^{(1)} \leftarrow \left|\mathbf{L} * f_{i+1}^d\right|$. \\
2. For each coarse scale $k \in \{2,4\}$, compute
$\tilde{f}_{i+1}^{(k)} \leftarrow \operatorname{MaxPool}(f_{i+1}^d; k)$ and
$E^{(k)} \leftarrow \operatorname{Upsample}\!\left(\left|\mathbf{L} * \tilde{f}_{i+1}^{(k)}\right|\right)$. \\
3. Aggregate cross-scale edge cues:
$f_i^{\mathrm{edge}} \leftarrow \operatorname{Norm}\!\left(\frac{1}{3}\sum_{k \in \{1,2,4\}} E^{(k)}\right)$. \\
\textbf{Phase II: Spatial modulation} \\
4. Global and local spatial responses:
$f_i^{g} \leftarrow \operatorname{GE}(f_i^e)$, \;
$f_i^{\mathrm{loc}} \leftarrow \operatorname{LE}(f_i^e)$. \\
5. Edge-modulated attention map:
$f_i^{m} \leftarrow (f_i^{g} + f_i^{\mathrm{loc}}) \odot f_i^{\mathrm{edge}}$. \\
\textbf{Phase III: Feature refinement and channel recalibration} \\
6. Concatenation and residual refinement:
$f_i^c \leftarrow \operatorname{Conv}_{3\times3}\!\big(\operatorname{Concat}(f_i^e, f_i^{g}, f_i^{\mathrm{loc}}, f_i^{m})\big) + f_i^e$. \\
7. Channel recalibration via ECA:
$f_i^{a} \leftarrow f_i^c \otimes \sigma\!\big(\operatorname{Conv1D}_{\kappa}(\operatorname{GAP}(f_i^c))\big)$, \;
$\kappa = \psi(C_i)$. \\
\bottomrule
\end{tabular}
\end{table}

\begin{table}[t]
    \centering
    \caption{Quantitative comparison of MambaMPD on the MADOS dataset(Excluding foundation model).\\ Note: U-Net is included here as one of the compared methods for overall benchmarking; the module-wise ablation on the core Mamba architecture is reported in Table 6.}
    % 不用fm,放上marinext
    \label{tab:mados}
    \setlength{\tabcolsep}{4.5pt}
    \renewcommand{\arraystretch}{0.95}
    \begin{tabular}{lccc}
        \toprule
        \textbf{Model} & \textbf{F1 (\%, \(\uparrow\))} & \textbf{mIoU (\%, \(\uparrow\))} & \textbf{OA (\%, \(\uparrow\))} \\
        \midrule
        RF\cite{rf}                     & 56.6 & 43.9 & 67.1 \\
        RF++\cite{RF++}                 & 64.4 & 52.4 & 83.8 \\
        U-Net\cite{ronneberger2015u}    & 63.8 & 51.0 & 82.9 \\
        SegNeXt\cite{segnext}           & 60.6 & 49.2 & \textbf{86.6} \\
        MariNeXt \cite{kikaki2024detecting}      & 70.6 & 59.2 & 81.6 \\
        OSDMamba\cite{OSDMamba}      & 71.2 & 68.1 & 82.3 \\
        MambaMPD (ours)                 & \textbf{74.8} & \textbf{69.8} & 83.1 \\
        \bottomrule
    \end{tabular}
\end{table}

\begin{table}[t]
    \centering
    \caption{Quantitative comparison of our MambaMPD model against recent remote sensing foundation models on the MADOS dataset.}
    \label{tab:mados_miou}
    \setlength{\tabcolsep}{4pt}
    \renewcommand{\arraystretch}{0.95}
    \begin{tabular}{lc}
        \toprule
        \textbf{Model} & \textbf{mIoU (\%, \(\uparrow\))} \\
        \midrule
        % RF\cite{rf}                 & 43.9 \\
        % RF++\cite{RF++}             & 52.4 \\
        % U-Net\cite{ronneberger2015u}& 51.0 \\
        % SegNext\cite{segnext}       & 49.2 \\
        % CROMA\cite{fuller2023croma}    & 67.6 \\
        DOFA\cite{xiong2025neuralplasticityinspiredmultimodalfoundation} & 59.58 \\
        GFM-Swin\cite{lu2025vision} & 64.7 \\
        prithvi\cite{szwarcman2024prithvi} & 49.9 \\
        RemoteCLIP\cite{liu2024remoteclip} & 60.0 \\
        SatlasNet\cite{bastani2023satlaspretrain} & 55.9 \\
        Scale-MAE\cite{reed2023scale} & 57.3 \\
        SpectralGPT\cite{hong2023spectralgpt} & 57.9 \\
        SSL4EO-S12-MoCo\cite{wang2023ssl4eo} & 51.8 \\
        SSL4EO-S12-DINO\cite{wang2023ssl4eo} & 49.4 \\
        SSL4EO-S12-MAE\cite{wang2023ssl4eo} & 49.9 \\
        SSL4EO-S12-Data2Vec\cite{wang2023ssl4eo} & 44.4 \\
        TerraMind-B (TerraMesh)\cite{blumenstiel2025terramesh} & 69.5 \\
        PANGAEA(U-Net)\cite{marsocci2025pangaeaglobalinclusivebenchmark} & 57.8 \\
        PANGAEA(VITB-16)\cite{marsocci2025pangaeaglobalinclusivebenchmark} & 48.2 \\
        \textbf{MambaMPD (ours)} & \textbf{69.8} \\
        \bottomrule
    \end{tabular}
\end{table}
\section{Experiments}

\subsection{Datasets}
1) \textbf{MADOS Dataset} \cite{kikaki2024detecting}:
The MADOS (Marine Debris and Oil Spill) dataset is a globally distributed benchmark dataset specifically designed for detecting marine pollution, including oil spills and marine debris. It contains 174 multispectral Sentinel-2 (S2) satellite images, captured between 2015 and 2022, with approximately 1.5 million pixels annotated. The dataset covers 15 different thematic categories, such as oil spills, marine debris, ships, phytoplankton, turbid waters, etc. The S2 images in the MADOS dataset cover global coastal waters with spatial resolutions of 10 meters, 20 meters, and 60 meters, and a revisit cycle of about 5 days. During the data processing, ACOLITE atmospheric correction~\cite{vanhellemont2018atmospheric} was used to extract Rayleigh reflectance, and annotations were provided by multiple experts, with a focus on labelling major pollutants like oil spills and marine debris. 
Additionally, the dataset exhibits significant diversity, covering various geographic distributions and environmental conditions, capturing different weather and ocean states. We adopt the official scene-level data split 
provided by Kikaki et al.~\cite{kikaki2024detecting}, where 
training and test samples correspond to geographically 
distinct Sentinel-2 acquisitions, preventing spatial leakage 
between partitions.

2) \textbf{M4D Dataset} \cite{krestenitis2019oil}:
The M4D dataset was developed for marine oil spill detection research. The dataset consists of images extracted from satellite synthetic aperture radar (SAR) data, covering oil spills and other related semantic categories, with corresponding ground truth masks and labels~\cite{krestenitis2019oil}. Specifically, the dataset comprises 1002 training images and 110 test images, each with corresponding labels. A total of 5 semantic classes are annotated, including: sea surface, oil spill, look-alike, ship, and land. We use the fixed 1002/110 train/test split of
Krestenitis et al.~\cite{krestenitis2019oil}; training and
test images come from separate SAR acquisitions with no
spatial overlap. We hold out 5\% of the training set for
validation.

\subsection{Implementation Details}
Our network is built using the PyTorch framework. In the M4D dataset, the training template we used follows the design in HTSM~\cite{satyanarayana2023oil}. We trained the model using an SGD optimizer with a momentum of 0.9 and a weight decay of 1e-4. Additionally, we set the initial learning rate to 0.001 and employed a ``Poly'' decay strategy. All experiments were implemented on an NVIDIA 4070Ti GPU. The batch size was set to 8, with a maximum of 100 epochs. Similarly, in the MADOS dataset, we used the official training framework provided by~\cite{kikaki2024detecting}. For classical CNN methods, the backbone networks employed during the encoder stage were pretrained on ImageNet. Following OSDMamba~\cite{OSDMamba}, we use a hybrid objective that combines Focal Loss and Jaccard Loss, formulated as \( L = -\alpha (1 - p_t)^{\gamma} \log(p_t) + \left(1 - \frac{\sum_i TP_i}{\sum_i TP_i + \sum_i FP_i + \sum_i FN_i}\right) \), where \( \alpha \) balances class frequencies, \( p_t \) is the predicted probability of the true class, \( \gamma \) modulates the emphasis on hard examples, and \( TP_i, FP_i, FN_i \) denote pixel-wise statistics for class \( i \). The model was initialized with pre-trained weights from ImageNet~\cite{imagenet}. A batch size of 4 was used, and the model was trained for 100 epochs across all stages. Deep supervision is applied across decoding stages, and the final training objective sums the hybrid loss with all auxiliary prediction losses using predefined layer-wise weights. All methods are compared on the official
train/test split of each dataset with identical metric computation. The
CNN- and Transformer-based results in Tables~2 and~4 are obtained on
these official splits, with each model trained end-to-end from an
ImageNet-pretrained backbone under its original configuration. The
foundation-model results in Table~3 follow the PANGAEA evaluation
framework~\cite{marsocci2025pangaeaglobalinclusivebenchmark}: each model is initialised with its publicly released
weights obtained by large-scale pre-training on Earth observation data
(masked autoencoding, contrastive learning, or supervised pre-training),
its encoder is frozen as a feature extractor, and a UPerNet decoder fed
with four intermediate feature levels is trained on the official MADOS
training split, using the same decoder, optimiser, schedule, and band
adaptation for all models. MambaMPD is trained end-to-end from
ImageNet-pretrained weights on the same splits.

\subsection{Evaluation Metrics}
For the M4D dataset, we use the mean Intersection over Union (mIoU) score over the union to evaluate the model performance. In the MADOS dataset, both mIoU and average F1 (Ave.F1) scores are used to evaluate the model performance. These two evaluation metrics are based on the confusion matrix, which contains four components: True Positives (TP), False Positives (FP), True Negatives (TN), and False Negatives (FN). For each class, IoU is defined as the ratio of the intersection and union of the predicted and ground truth values. In addition, we report macro-averaged Precision, 
Recall, and F1-score on the M4D dataset 
(Table~\ref{tab:m4d_extra_metrics}) to provide a more 
comprehensive assessment of model performance.

\begin{figure*}
    \centering
    \includegraphics[width=5.5in]{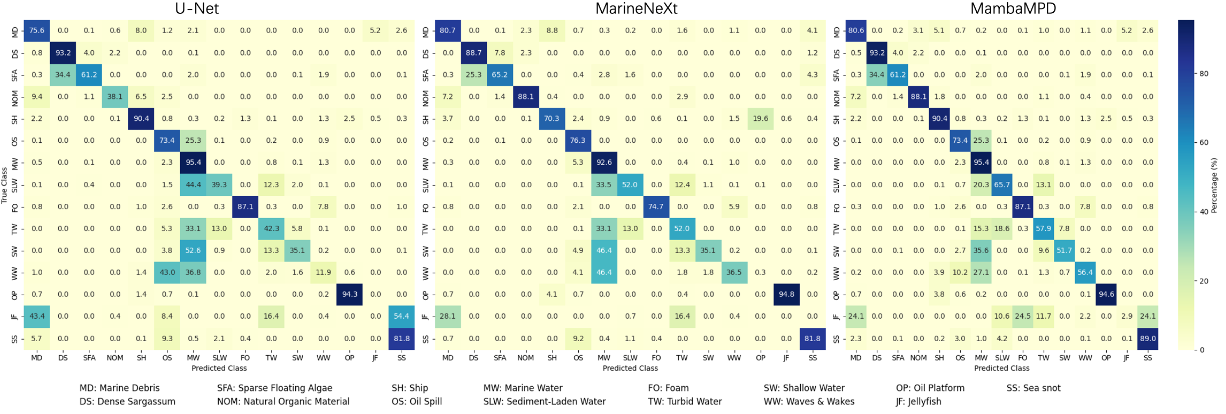}
    \caption{Comparison of confusion matrices on the MADOS dataset.}
    \label{fig:heatmap}
\end{figure*}

\begin{table*}[t]
    \centering
    \caption{Quantitative performance comparison on the M4D dataset.}
    \label{tab:m4d}
    \small
    \setlength{\tabcolsep}{3.5pt}
    \renewcommand{\arraystretch}{0.95}
    \resizebox{0.96\textwidth}{!}{
    \begin{tabular}{lccccccc}
        \toprule
        \textbf{Model} &
        \textbf{Sea Surface(\%)} &
        \textbf{Oil Spill(\%)} &
        \textbf{Look alike(\%)} &
        \textbf{Ship(\%)} &
        \textbf{Land(\%)} &
        \textbf{mIoU(\%)} &
        \textbf{FPS} \\
        \midrule
        Unet\cite{ronneberger2015u}                  & 93.90 & 53.79 & 39.55 & 44.93 & 92.68 & 64.97 & 52.16 \\
        LinkNet\cite{Linknet}            & 94.99 & 51.53 & 43.24 & 40.23 & 93.97 & 64.79 & 66.45 \\
        PSPNet\cite{PSPNet}              & 92.78 & 40.10 & 33.79 & 24.42 & 86.90 & 55.60 & 25.23 \\
        DeepLabv2\cite{Deeplabv2}        & 94.09 & 25.57 & 40.30 & 11.41 & 74.99 & 49.27 & 15.62 \\
        DeepLabv2 (msc)\cite{Deeplabv2}  & 95.39 & 49.53 & 49.28 & 31.26 & 88.65 & 62.83 & 16.14 \\
        DeepLabv3+\cite{deeplabv3}       & 96.43 & 53.38 & 55.40 & 27.63 & 92.44 & 65.06 & 55.10 \\
        YOLOv8-SAM\cite{wu2024compositional} & 94.34 & 41.84 & 48.15 & 52.48 & 87.65 & 64.89 & -- \\
        SAM-OIL\cite{wu2024compositional}     & 96.05 & 51.60 & 55.60 & \textbf{52.55} & 91.81 & 69.52 & -- \\
        TransOilSeg\cite{chai2025transoilseg} & \textbf{97.02} & 61.38 & \textbf{62.41} & 33.49 & \textbf{94.39} & 69.74 & 8.86 \\
        OSDMamba\cite{OSDMamba}          & 96.47 & 65.59 & 47.57 & 46.85 & 94.76 & 70.25 & 10.00 \\
        \midrule
        \textbf{MambaMPD (ours)}         & 96.30 & \textbf{68.20} & 43.60 & 52.36 & 93.79 & \textbf{70.85} & 15.19 \\
        \bottomrule
    \end{tabular}
    }
\end{table*}

\begin{table}[t]
\centering
\caption{Additional evaluation metrics on the M4D 
dataset. Precision, Recall, and F1-score are computed as the 
macro-average across all five classes.}
\label{tab:m4d_extra_metrics}
\small
\begin{tabular}{lcccc}
\toprule
Model & Precision (\%) & Recall (\%) & F1 (\%) & mIoU (\%) \\
\midrule
U-Net         & 77.80 & 74.75 & 75.50 & 64.97 \\
DeepLabV3+    & 72.72 & 77.05 & 74.57 & 65.06 \\
TransOilSeg   & 79.77 & 78.33 & 78.99 & 69.74 \\
OSDMamba      & 80.13 & 78.21 & 77.35 & 70.25 \\
MambaMPD (ours) & 81.02 & 79.41 & 78.18 & 70.85 \\
\bottomrule
\end{tabular}
\end{table}
\subsection{Comparison With State-of-the-Art}

\subsubsection{Quantitative Results on MADOS Dataset}
% We evaluated and compared our MambaMPD with RF~\cite{rf}, RF++~\cite{RF++}, U-Net~\cite{ronneberger2015u}, SegNeXt~\cite{segnext}, and Mari-Next~\cite{kikaki2024detecting} on the MADOS dataset. 
We evaluated and compared our MambaMPD with RF~\cite{rf}, RF++~\cite{RF++}, U-Net~\cite{ronneberger2015u}, SegNeXt~\cite{segnext}, MariNeXt (reproduced)~\cite{kikaki2024detecting}, 
% and a total of 18 representative baseline models~\cite{fuller2023croma,xiong2025neuralplasticityinspiredmultimodalfoundation,lu2025vision,szwarcman2024prithvi,liu2024remoteclip,bastani2023satlaspretrain,reed2023scale,hong2023spectralgpt,wang2023ssl4eo,blumenstiel2025terramesh,marsocci2025pangaeaglobalinclusivebenchmark} on the MADOS dataset. 
and approaches involving fine-tuning of remote sensing foundation models, such as CROMA, DOFA, GFM-Swin, Prithvi, RemoteCLIP, SatlasNet, Scale-MAE, SpectralGPT, SSL4EO, TerraMind-B, and PANGAEA~\cite{fuller2023croma,xiong2025neuralplasticityinspiredmultimodalfoundation,lu2025vision,szwarcman2024prithvi,liu2024remoteclip,bastani2023satlaspretrain,reed2023scale,hong2023spectralgpt,wang2023ssl4eo,blumenstiel2025terramesh,marsocci2025pangaeaglobalinclusivebenchmark}.
Table~\ref{tab:mados} and Table~\ref{tab:mados_miou} report the segmentation performance of each method on the MADOS dataset, further demonstrating the effectiveness of our proposed MambaMPD. Our MambaMPD achieves an F1 score of 74.8\% and an mIoU of 69.8\%, outperforming the other compared methods. 
Compared to classical ensemble methods such as RF (F1: 56.6\%, mIoU: 43.9\%) and RF++ (F1: 64.4\%, mIoU: 52.4\%), MambaMPD shows substantial improvements, particularly in terms of segmentation quality. When compared to U-Net~\cite{ronneberger2015u} and SegNeXt~\cite{segnext}, which are popular CNN-based architectures, MambaMPD delivers significant gains of 18.8\% and 20.6\% in mIoU, respectively. Moreover, it also outperforms MariNeXt~\cite{kikaki2024detecting}, a multi-branch CNN designed for multi-type marine pollutant recognition, by a notable margin of 10.6\% in mIoU and 4.2\% in F1 score. These results clearly demonstrate the superior capability of MambaMPD in handling complex and heterogeneous marine scenes. Its strong performance across all three metrics reflects not only accurate segmentation but also robustness in classifying fine-grained and ambiguous targets under real-world oceanic conditions.

Furthermore, we computed the confusion matrix obtained by applying MambaMPD on the MADOS test set, as shown in Fig.~\ref{fig:heatmap}. Our model shows better performance compared with other models. Although the number of test pixels for oil spill categories is relatively low compared to others, our model still achieves good segmentation performance. Specifically, the F1 scores for small-object classes consistently exceed 70\%, reflecting the model's robustness in addressing class imbalance. Notably, the accuracy for the Ship class reaches 90.4\%, indicating that our model exhibits strong discriminative capability for key categories, effectively reducing misclassification.

\begin{figure*}[t]
    \centering
    \includegraphics[width=5.5in]{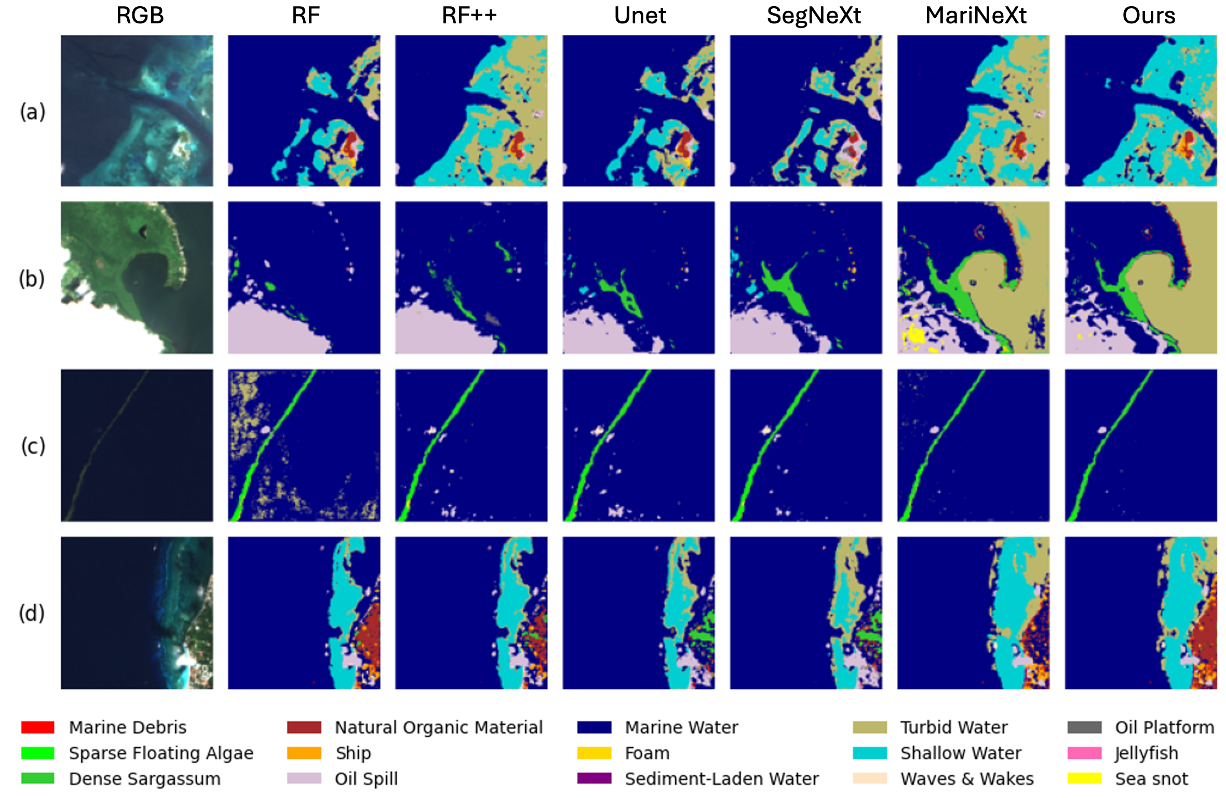}
    \caption{Qualitative performance comparison on selected test samples of the MADOS dataset.}
    \label{fig:madosvis}
\end{figure*}

\begin{figure*}[t]
    \centering
    \includegraphics[width=5.5in]{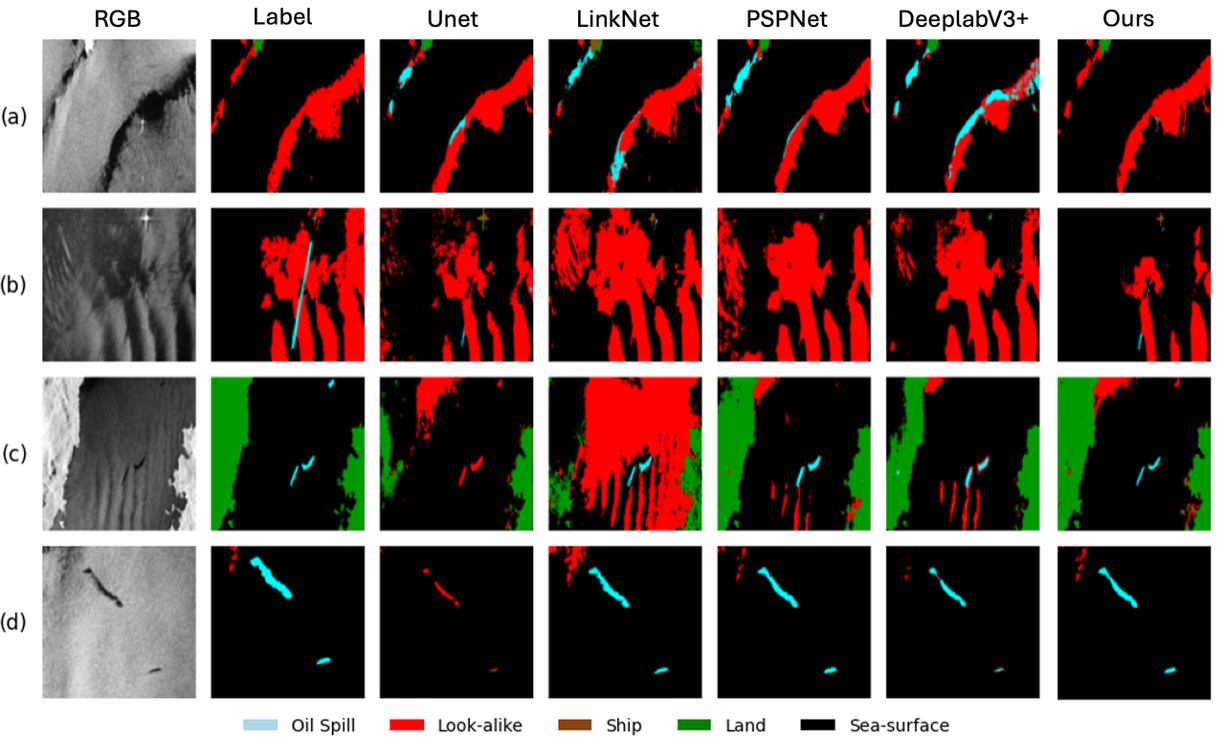}
    \caption{Qualitative performance comparison on the selected test samples of the M4D dataset.}
    \label{fig:m4dvis}
\end{figure*}

\subsubsection{Quantitative Results on M4D Dataset}

We further evaluated and compared the proposed MambaMPD with several existing methods, including Unet~\cite{ronneberger2015u}, LinkNet~\cite{Linknet}, PSPNet~\cite{PSPNet}, Deeplabv2~\cite{Deeplabv2}, Deeplabv2 (msc)~\cite{Deeplabv2}, Deeplabv3+~\cite{deeplabv3}, YOLOv8-SAM~\cite{wu2024compositional}, SAM-OIL~\cite{wu2024compositional} and TransOilSeg~\cite{chai2025transoilseg} on the M4D dataset. 

Table~\ref{tab:m4d} demonstrates that our MambaMPD outperforms other methods in both oil spill IoU and total mIoU. 
We further compared MambaMPD with models built upon the Segment Anything Model (SAM), specifically YOLO-SAM and SAM-OIL, which leverage prompt-based or detection-driven strategies to address oil spill segmentation. While these methods benefit from SAM’s general-purpose segmentation capability, they lack dedicated mechanisms for modelling structured context or preserving fine-grained boundaries. In contrast, our MambaMPD leverages the selective scanning mechanism of the Mamba architecture to explicitly capture long-range dependencies and improve contextual understanding across sparse oceanic targets.
In addition, we compared MambaMPD with mainstream CNN-based segmentation networks. DeepLabV3+~\cite{deeplabv3} utilises atrous spatial pyramid pooling (ASPP) to encode multi-scale context, while U-Net employs hierarchical skip connections to preserve spatial resolution. Although both achieve reasonable performance, they are constrained by local receptive fields and struggle with long-range semantic modeling. MambaMPD significantly outperforms these baselines, especially in the oil spill category, where it achieves a remarkable 14.82\% improvement in Oil Spill IoU over DeepLabV3+ and a 26.36\% gain compared to YOLO-SAM. These results clearly demonstrate the advantage of combining state-space modeling with frequency- and edge-aware modules for structure-sensitive marine pollutant detection.

OSDMamba shares the same VSS backbone
family and training objective, which makes it the most direct reference
for assessing the proposed modules. MambaMPD improves Oil Spill IoU from
65.59\% to 68.20\% on M4D and F1 from 71.2\% to 74.8\% on MADOS. The
ablation study attributes these gains to the two modules: FAA alone
raises Oil Spill IoU by +5.29\% over the Mamba baseline (Table~6), and
replacing EGA with SE or CBAM attention degrades all boundary-sensitive
classes (Table~7). The improvements therefore stem from the
frequency-aware and edge-guided designs rather than from model capacity.

To complement the IoU-based evaluation,
Table~\ref{tab:m4d_extra_metrics} further reports macro-averaged
Precision, Recall, and F1-score. MambaMPD attains the highest Precision
(81.02\%) and Recall (79.41\%); its F1-score (78.18\%) is second to
TransOilSeg (78.99\%), while its mIoU remains the highest among all
compared methods.

\subsubsection{Qualitative Analysis}
Additional examples on the MADOS test set are shown in Fig.~\ref{fig:madosvis}.  Neither a classical Random Forest classifier nor mainstream encoder–decoder architectures (e.g., U-Net) nor the recent MariNeXt network succeed in recovering satisfactory edge or texture detail from the multi-spectral data: land–sea boundaries are frequently mis-labelled, and objects with blurred contours are often missed.  In contrast, MambaMPD faithfully reconstructs subtle high-frequency structures, such as ship wakes and thin pollutant filaments, while simultaneously suppressing noise.  This superior visual performance underscores the model’s capacity to exploit edge-aware priors and long-range context, enabling the reconstruction of details that closely match the true spatial distribution of marine pollutants.
% Across every scene, MambaMPD delivers the most accurate and visually coherent delineation of pollution targets
A visual comparison with all state-of-the-art baselines is presented in Fig.~\ref{fig:m4dvis}. In the majority of evaluated scenes, MambaMPD 
delivers more accurate and visually coherent 
delineation of pollution targets, generally outperforming both traditional 
convolutional and modern Transformer-based encoders.
In rows (a) and (b) of Fig.~\ref{fig:m4dvis}, conventional encoder–decoder networks such as PSPNet misclassify large portions of the slick, reflecting their limited capability to resolve fine-scale boundaries.  Although earlier deep learning methods better capture overall shape and edge information, they remain susceptible to texture-induced artefacts.  By contrast, MambaMPD obtains sharp, artefact-free borders that follow the ground-truth contour with high fidelity, demonstrating the benefit of its wavelet-enhanced Mamba encoder and edge-guided attention. In the cluttered scene highlighted in Fig.~\ref{fig:m4dvis}, all competing models are distracted by background objects and consequently yield false positives. Owing to its stronger discriminative power in visually ambiguous and cluttered scenes, MambaMPD more reliably isolates the pollutant region correctly and preserves background integrity.  These observations suggest that our framework can 
exploit richer structural priors, thereby improving recognition 
robustness in the tested marine environments.

\subsubsection{Failure Case Analysis}

\begin{figure*}[t]
    \centering
    \includegraphics[width=5.5in]{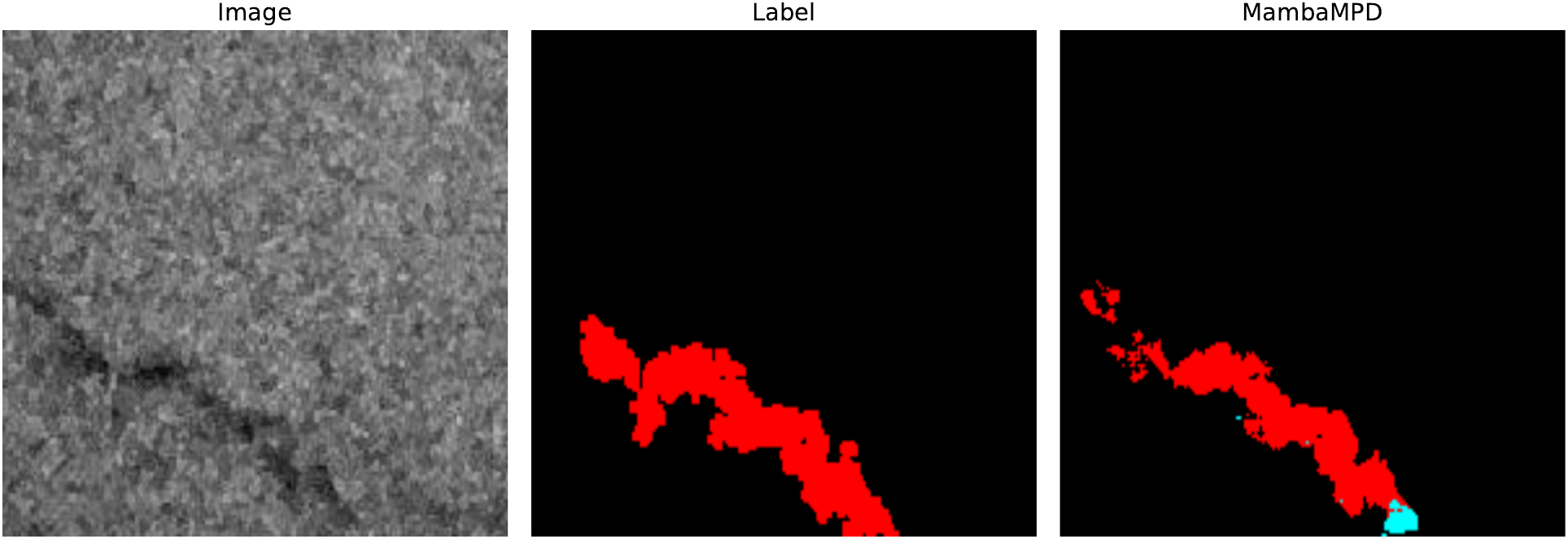}
    \caption{Representative failure case on the M4D test set. SAR
  image containing an elongated dark formation; Label, in
  which the entire formation is annotated as Look-alike; MambaMPD
  prediction. Most of the formation is correctly identified, but its
  darkest and most homogeneous segment is labelled as Oil
  Spill: locally, this segment carries the low-backscatter signature
  that the frequency- and edge-enhanced features amplify.}
    \label{fig:failure}
\end{figure*}

MambaMPD achieves the highest overall mIoU but drops to 43.60\% on the
Look-alike class, well behind TransOilSeg and SAM-OIL. The frequency-enhanced and edge-guided features push the
model to detect dark-patch structures aggressively: this raises Oil
Spill IoU to 68.20\%, the best among all methods, but also misclassifies
look-alike regions such as biogenic films and low-wind areas as Oil
Spill. Fig.~\ref{fig:failure} shows a
representative case. The darkest and most homogeneous segment of an
elongated look-alike formation, where the low-backscatter signature of
oil is locally strongest, is predicted as Oil Spill; the remainder of
the formation is correctly identified. Separating such segments requires
context beyond local backscatter, such as shape regularity or wind
conditions, which the current features do not capture.

TransOilSeg exhibits the reverse trade-off, leading on Look-alike but
falling to 61.38\% on Oil Spill, so the two classes currently sit on a
recall--precision frontier rather than being jointly solved by either
design. Future work will explore
class-aware contrastive learning and look-alike-specific auxiliary
losses to better separate these visually similar categories.

\begin{table*}[t]
    \centering
    \caption{Ablation study of the proposed modules on the M4D dataset.}
    \label{tab:ablation}

    \setlength{\tabcolsep}{4.5pt}
    \renewcommand{\arraystretch}{1.15}
    \resizebox{\textwidth}{!}{%
    \begin{tabular}{ccc|ccccc|cc}
        \toprule
        \rule{0pt}{2.5ex}
        \textbf{FAA} & \textbf{SER} & \textbf{EGA} &
        \textbf{SeaSurface (\%, $\uparrow$)} &
        \textbf{Oil Spill (\%, $\uparrow$)} &
        \textbf{Look-alike (\%, $\uparrow$)} &
        \textbf{Ship (\%, $\uparrow$)} &
        \textbf{Land (\%, $\uparrow$)} &
        \textbf{MIoU (\%, $\uparrow$)} &
        \textbf{Acc (\%, $\uparrow$)} \\
        \midrule
        & & & 94.93 & 61.68 & 45.14 & 45.23 & 91.01 & 67.51 & 95.73 \\
         % &  &  & 94.93 & 61.68 & 45.14 & 45.23 & 91.01 & 67.51 & 96.13 \\
        \checkmark &  &  & 95.43 & 66.97 & 46.09 & 37.58 & 91.82 & 67.58 & 96.15 \\
         & \checkmark &  & 95.58 & 66.70 & 45.51 & 37.55 & 93.26 & 67.72 & 96.24 \\
         &  & \checkmark & 96.40 & 65.46 & 46.70 & 38.36 & 94.59 & 68.30 & 97.00 \\
        \checkmark & \checkmark &  & 96.29 & 65.38 & 47.62 & 43.29 & 94.14 & 69.34 & 96.93 \\
         & \checkmark & \checkmark & 95.73 & 67.36 & 47.38 & 41.88 & 93.18 & 69.11 & 96.49 \\
        \checkmark &  & \checkmark & 96.27 & 67.04 & 45.16 & 43.36 & 95.57 & 69.48 & 95.39 \\
        \checkmark & \checkmark & \checkmark & 96.30 & \textbf{68.20} & 43.60 & 52.36 & 93.79 & \textbf{70.85} & \textbf{97.21} \\
        \bottomrule
    \end{tabular}%
    }
\end{table*}

\subsection{Ablation Study}

To thoroughly evaluate the effectiveness of each proposed module in our architecture, we conduct extensive ablation experiments on the M4D dataset. 
The baseline model is a MambaU-Net consisting of 
a VSS encoder with \{2,\,2,\,9,\,2\} blocks and a standard 
UnetrUpBlock decoder, without any of the proposed modules (FAA, 
SER, or EGA). Each module is then individually or jointly added 
to this Mamba-based baseline to assess its contribution. The results are shown in Table~\ref{tab:ablation}. The investigated modules include: the Frequency-Aware Augmentation Module (FAA) for frequency-aware feature extraction, the SE-ResDecoder (SER) for semantic enrichment, and the Edge-Guided Attention (EGA) for fine boundary preservation. We also analyse the role of their combinations and the cumulative effect when all modules are used together.

\subsubsection{Effect of Single Modules}

Without applying FAA, SER, or EGA, the baseline MambaU-Net achieves an Oil IoU of 61.68\% and an mIoU of 67.51\%. This setting serves as the reference point for evaluating the contribution of each individual module.

When only the Frequency-Aware Augmentation (FAA) module is added, the Oil IoU improves significantly from 61.68\% to 66.97\%, and the mIoU rises from 67.51\% to 67.58\%. This confirms that FAA enhances the model’s ability to extract discriminative multi-frequency representations, which are crucial for capturing subtle structural, textural, and contextual differences between marine pollutants and ocean backgrounds.

Introducing only the SE-ResDecoder (SER) module yields an Oil IoU of 66.70\% and an mIoU of 67.72\%. The improvement compared with the baseline demonstrates that SER effectively enriches geometric and structural cues, enabling better discrimination of irregular pollutant shapes within SAR imagery.

Using only the Edge-Guided Attention (EGA) module increases Oil IoU to 65.46\% and mIoU to 68.30\%, representing the strongest single-module enhancement. This verifies that explicitly modeling boundary priors greatly benefits marine pollution segmentation by mitigating the edge-blurring tendency of SSM-based architectures.
\subsubsection{Effect of Dual Module Combinations}

We then evaluate the synergy between two modules. The combination of FAA and SER improves the segmentation accuracy significantly, yielding an oil spill IoU of 65.38\% and an mIoU of 69.34\%. This suggests that frequency-aware and semantic-enhancing modules complement each other by capturing global context while preserving class-specific features.
Combining FAA and EGA (without SER) also achieves strong results, with an oil spill IoU of 67.04\% and mIoU of 69.48\%, indicating that frequency and boundary information together help the model to suppress noise and highlight transition regions. On the other hand, the SER + EGA setting (without FAA) results in 67.36\% oil spill IoU and 69.11\% mIoU, which is inferior to the configurations involving FAA, highlighting the foundational role of frequency-aware encoding in marine pollution scenarios.

\subsubsection{Full Model with All Modules}

Finally, when all three modules (FAA + SER + EGA) are integrated, the model achieves the best performance across nearly all categories. Specifically, it obtains an oil spill IoU of 68.20\%, an overall mIoU of 70.85\%, and a pixel-level accuracy of 97.21\%. This configuration shows superior performance not only in the oil spill class but also across the Sea Surface and Ship classes, suggesting that each module contributes uniquely and effectively to the final representation.
These results clearly demonstrate that the three modules provide complementary advantages: FAA improves global frequency perception, SER enhances semantic discriminability for rare categories, and EGA refines boundaries and edge structures. Their joint effect enables the model to capture both global context and fine-grained details, which are critical for robust oil spill segmentation under real-world, complex marine conditions.

\begin{table}
    \centering
    \small
    \caption{Sensitivity of decoder attention strategies on the M4D dataset.
    Backbone (VSS $\{2,2,9,2\}$), FAA, and SE-ResDecoder remain fixed; only the
    attention module is replaced.}
    \label{tab:att_sensitivity}
    % \normalsize
    \setlength{\tabcolsep}{5.5pt}
    \renewcommand{\arraystretch}{1.08}
    \begin{tabular}{lccc}
        \toprule
        \textbf{Attention type} 
            & \textbf{Oil(\%)} 
            & \textbf{Look-alike(\%)}
            & \textbf{mIoU(\%)} \\
        \midrule
        None (no att.)        & 65.38 & 47.62 & 69.34 \\
        SE                    & 69.44 & 41.82 & 69.88 \\
        CBAM                  & 69.87 & 42.10 & 70.12 \\
        \textbf{EGA (ours)}   & \textbf{68.20} & \textbf{43.60} & \textbf{70.85} \\
        \bottomrule
    \end{tabular}
\end{table}

\section{Further Analysis}
To more comprehensively assess the robustness and efficiency of the proposed
MambaMPD framework, we conduct additional analyses from three complementary perspectives: architectural sensitivity to wavelet configurations in the FAA
module, decoder attention strategies, and overall computational complexity.
These studies provide deeper insights into why MambaMPD exhibits strong
boundary modelling capability while maintaining its lightweight nature.

\begin{table}
    \centering
    \caption{Sensitivity of the FAA module to different wavelet bases.
    The decomposition 
    level is fixed to $J=2$. Results are reported on both 
    M4D and MADOS to verify cross-modality robustness.}
    \label{tab:wavelet_basis}
    \small
    \setlength{\tabcolsep}{4pt}
    \renewcommand{\arraystretch}{1.06}
    \begin{tabular}{l ccc ccc}
        \toprule
        & \multicolumn{3}{c}{\textbf{M4D}} 
        & \multicolumn{3}{c}{\textbf{MADOS}} \\
        \midrule
        \textbf{Wavelet basis} 
        & \textbf{OA} & \textbf{mIoU} & \textbf{Oil IoU} 
        & \textbf{OA} & \textbf{mIoU} 
        & \textbf{F1} \\
        \midrule
        Haar               & 97.11 & 70.78 & 67.91
& 83.22 & 69.73 
                           & 74.15 \\
        Daubechies-2 (db2) & 97.64 & 70.53 & 67.01 
                           & 82.75 & 69.11 
                           & 74.69 \\
        Daubechies-4 (db4) & 97.65 & 70.27 & 67.69 
                           & 83.34 & 69.60 
                           & 74.71 \\
        \bottomrule
    \end{tabular}
\end{table}

% ============================================================
% Table 8: Decomposition level sensitivity — M4D + MADOS
% ============================================================

\begin{table}
\centering
\caption{Sensitivity of the FAA module to the wavelet decomposition
level $J$. The wavelet basis is fixed to Haar.
Results are reported on both M4D and MADOS.}
\label{tab:wavelet_level}
\small
\setlength{\tabcolsep}{4pt}
\renewcommand{\arraystretch}{1.06}
\begin{tabular}{l ccc ccc}
\toprule
& \multicolumn{3}{c}{\textbf{M4D}}
& \multicolumn{3}{c}\textbf{MADOS} \\
\midrule
\textbf{Level $J$}
& \textbf{OA} & \textbf{mIoU} & \textbf{Oil IoU}
& \textbf{OA} & \textbf{mIoU}
& \textbf{F1} \\
\midrule
$J = 1$ & 97.63 & 70.16 & 67.44
& 83.17 & 68.93
& 74.68\\
$J = 2$ & 97.11 & 70.78 & 67.91
& 83.22 & 69.73
& 74.15  \\
$J = 3$ & 97.29 & 70.14 & 67.42
& 83.47 & 69.06
& 74.34 \\
\bottomrule
\end{tabular}
\end{table}

\subsection{Sensitivity to Wavelet Configuration in FAA}

The FAA module leverages multi–level discrete wavelet transform (DWT) to
extract high–frequency components that are critical for modelling thin oil
films and fragmented look–alike structures. As wavelet-based frequency
decomposition may, in principle, be influenced by the choice of wavelet
settings, we evaluate the sensitivity of FAA to two key hyperparameters:
(i)~the wavelet basis and (ii)~the decomposition level $J$. All experiments are
performed on the M4D dataset using the standard VSS backbone configuration
$\{2,2,9,2\}$ and the same training protocol as in the main study. As shown in Table~\ref{tab:wavelet_basis} and Table~\ref{tab:wavelet_level}.

\noindent \textbf{Wavelet basis.}
We fix the decomposition level to $J=2$ and examine three commonly used
wavelets: Haar, Daubechies-2 (db2), and Daubechies-4 (db4). As summarised in
Table~\ref{tab:wavelet_basis}, the resulting differences in OA, mIoU, and
oil–spill IoU are minimal, with mIoU fluctuating within $0.51\%$ and Oil IoU
within $0.90\%$. This demonstrates that the FAA is largely insensitive to the
specific wavelet family chosen, confirming that the gains of MambaMPD do not
depend on a finely tuned handcrafted basis.

\noindent \textbf{Decomposition level.}
Next, we vary the decomposition depth $J \in \{1,2,3\}$ while fixing the
wavelet basis to Haar (Table~\ref{tab:wavelet_level}). A shallow single-level
decomposition ($J=1$) results in a slightly reduced mIoU, indicating
that insufficient frequency resolution weakens multi–scale structure modelling.
Increasing the depth to $J=2$ yields the best performance, while
$J=3$ produces a marginal decline due to over-decomposition and loss of fine
spatial details. Importantly, the performance remains stable across all settings,
highlighting the robustness of FAA to reasonable changes in its wavelet
configuration.

\subsection{Sensitivity to Attention Strategies}

To assess whether the benefits of EGA depend on a particular attention form,
we compare EGA against three alternatives while keeping the VSS backbone, FAA,
and SE-ResDecoder fixed: (i)~no attention, (ii)~SE-based channel attention, and
(iii)~CBAM, a widely used spatial–channel attention module. As shown in
Table~\ref{tab:att_sensitivity}, adding attention generally improves performance
over the plain skip-connection baseline. However, EGA achieves the best overall mIoU, the most boundary-ambiguous class; although SE and CBAM attain slightly higher Oil IoU, EGA delivers the strongest overall performance by markedly improving this harder class. The improvements are particularly pronounced in boundary-sensitive categories, demonstrating that explicit integration of multi–scale edge cues provides more effective structural guidance than generic attention mechanisms.

Moreover, the overall variation across attention strategies remains small, confirming that MambaMPD is not overly sensitive to the
attention formulation. The superior performance of EGA arises from its
edge-guided reweighting rather than reliance on a fragile hyperparameter setup.

\begin{table*}
    \centering
    \caption{ Module-wise computational overhead analysis. Starting from the
baseline MambaU-Net, each proposed module is cumulatively added to
quantify its impact on parameters, FLOPs, and inference speed.}
\label{tab:complexity}
\small
\begin{tabular}{lccc}
\toprule
Model & Params (M) & FLOPs (G) & FPS \\
\midrule

Baseline MambaU-Net & 36.59 & 164.21 & 17.19 \\
\;+ FAA             & 36.89 & 185.61 & 16.17 \\
\;+ EGA             & 36.77 & 182.89 & 15.40 \\
\;+ FAA + EGA (Full)& 37.06 & 204.29 & 15.19 \\
\bottomrule
\end{tabular}
\end{table*}

\subsection{Computational Complexity}

To quantify the overhead introduced by each module, we measure trainable
parameters, FLOPs, and inference speed.  FLOPs are computed using
\texttt{ptflops} under a $1250 \times 650$ input resolution, and
throughput is measured as the average over 100 forward passes on an
NVIDIA RTX~4070Ti GPU.  Results are reported in
Table~\ref{tab:complexity}.

FAA adds 0.30\,M parameters and
21.40\,G FLOPs, reducing throughput from 17.19 to 16.17\,FPS.
EGA adds 0.18\,M parameters and 18.68\,G FLOPs.
With both modules active, the full MambaMPD model contains
37.06\,M parameters and 204.29\,G FLOPs---an increase of only
0.47\,M parameters and 40.08\,G FLOPs
over the baseline, while throughput remains at 15.19\,FPS.
By comparison, foundation-model-based approaches such as SAM-OIL
carry over 600\,M parameters, placing MambaMPD at roughly
$1/16$ of that capacity.
These figures confirm that the performance gains stem from the
structural design of the frequency-aware and edge-guided modules
rather than from scaling model size.

All current experiments use fixed-size patches; scaling
to sliding-window inference over full satellite swaths is left to
future work.

\section{Conclusion}

This paper presented MambaMPD, a segmentation framework for
marine pollution detection from remote sensing imagery.
MambaMPD builds on selective state-space modelling and introduces
two problem-driven modules: Frequency-Aware Augmentation (FAA) and
Edge-Guided Attention (EGA). Mamba offers linear-complexity
sequence modelling but is weak at capturing frequency structure
and preserving edges; FAA compensates through hierarchical wavelet
decomposition that survives spatial downsampling, while EGA injects
multi-scale Laplacian edge priors into the decoder, together
addressing the low contrast and boundary ambiguity of marine
pollutants in SAR and multispectral imagery.

Relative to prior work, MambaMPD differs from CNN and
Transformer-based detectors, which are limited by local receptive
fields, and from foundation-model-based approaches, which carry
hundreds of millions of parameters. To our knowledge, it is also
the first Mamba-based framework designed for multi-class marine
pollution detection across both SAR and multispectral modalities,
rather than binary oil-spill segmentation in a single modality.
This positioning translates into three practical advantages. First,
it attains the best mIoU among the compared methods on both
benchmarks, improving Oil Spill IoU by 6.82\% over TransOilSeg on
M4D and F1 by 3.6\% over OSDMamba on MADOS. Second, these gains come
at low cost: the full model adds only 0.47\,M parameters over the
baseline and runs faster than the strongest baselines, TransOilSeg
and OSDMamba, making it suitable for single-GPU deployment. Third,
its modality-agnostic design transfers between SAR and optical
pipelines without architectural changes. These properties make
MambaMPD a candidate for operational settings such as oil-spill
emergency response and routine coastal surveillance, and a
practical tool for maritime safety agencies, environmental
monitoring organisations, and coastal management authorities.

Several limitations remain. Look-alike discrimination remains imperfect:
the learned feature space does not yet separate pollutants from visually
similar backgrounds well enough. As analysed in Section 4.4.4, this reflects a deliberate emphasis on recall: a missed spill delays emergency response and its cost grows with time, whereas false alarms are filtered during routine human verification of alerts, so high recall on actual pollutants is prioritised in operational monitoring ~\cite{solberg2007oil,alpers2017oil}. Future work will address these gaps through joint SAR--optical fusion,
class-aware contrastive learning to suppress look-alike confusion, and
sliding-window inference over full satellite swaths. As suitable annotated
benchmarks become available, the framework can further be extended to
additional pollutant types, such as chemical contaminants and
microplastics, supporting more reliable and comprehensive marine
pollution assessment.

\section{Data Availability Statement}

The datasets used in this study are publicly available. The MADOS 
(Marine Debris and Oil Spill) dataset~\cite{mados_dataset} is openly available on Zenodo at 
\url{https://doi.org/10.5281/zenodo.10664073}. 
The Oil Spill Detection Dataset (M4D)~\cite{krestenitis2019oil} 
was originally provided by the Multimodal Data Fusion and Analytics 
Group at the Centre for Research and Technology Hellas (CERTH). 
All processed data used in the experiments of this study have been 
deposited on Zenodo and are publicly accessible at 
\url{https://doi.org/10.5281/zenodo.19386441}~\cite{m4d_dataset}.
The source code of the proposed MambaMPD framework is publicly 
available at \url{https://github.com/Multimodal-Intelligence-Lab-MIL/MambaMPD}~\cite{chen2026mambampd_code} under the CC0 v1.0 licence.

\section{Funding}
This work was supported in part by China Scholarship Council and in
part by the University of Exeter Ph.D. Scholarships. 
\section{Disclosure Statement}
No potential conflict of interest was reported by the author(s).

\renewcommand{\doi}[1]{doi: \href{https://doi.org/#1}{\nolinkurl{#1}}}% interact.cls defines \doi as a front-matter setter, which swallowed bibliography DOIs
\bibliographystyle{unsrtnat}
\bibliography{reference}
\end{document}